\documentclass[letterpaper, 10 pt, conference]{ieeeconf}  

\IEEEoverridecommandlockouts

\usepackage[T1]{fontenc} 
\usepackage{enumerate}
\usepackage{amsfonts}
\usepackage[colorinlistoftodos]{todonotes}
\usepackage[font=small,skip=12pt]{caption}
\usepackage{float}
\usepackage{subcaption}
\usepackage{bm}
\usepackage{booktabs}
\makeatletter
\let\NAT@parse\undefined
\makeatother
\usepackage{hyperref}
\usepackage{verbatim}
\usepackage{comment}
\usepackage{amsmath}
\usepackage{multirow}
\usepackage[noadjust]{cite}

\usepackage{mathtools}
\usepackage{dsfont}
\usepackage{tipa}
\usepackage{textcomp}
\usepackage[normalem]{ulem}
\DeclareMathAlphabet{\mathpzc}{OT1}{pzc}{m}{it}

\hypersetup{
    colorlinks=true,
    linkcolor=red,
    filecolor=magenta,      
    urlcolor=magenta,
    pdftitle={Overleaf Example},
    pdfpagemode=FullScreen,
    }
\usepackage{graphicx}
\usepackage{xcolor}
\usepackage[ruled,vlined,linesnumbered]{algorithm2e}

\SetCommentSty{mycommfont}
\newcommand{\myred}[1]{\textcolor{black}{#1}}

\title{\LARGE \bf
Hierarchical DLO Routing with Reinforcement Learning and In-Context Vision-Language Models}

\author{Mingen Li$^{1}$, Houjian Yu$^{1}$, Yixuan Huang$^{2}$, Youngjin Hong$^{1}$, Hantao Ye$^{1}$ and Changhyun Choi$^1$
\thanks{$^1$The authors are with the 
Department of Electrical and Computer Engineering, University of Minnesota, Minneapolis, MN 55455 {\tt\small \{li002852, yu000487, hong0745, ye000310, cchoi\}@umn.edu}}
\thanks{$^2$The author is with Princeton University, Princeton, NJ 08544 {\tt\small yh1542@princeton.edu}}%
}

\begin{document}

\maketitle
\thispagestyle{empty}
\pagestyle{empty}

\begin{abstract}
Long-horizon routing tasks of deformable linear objects (DLOs), such as cables and ropes, are common in industrial assembly lines and everyday life. 
These tasks are particularly challenging because they require robots to manipulate DLO with long-horizon planning and reliable skill execution. Successfully completing such tasks demands adapting to their nonlinear dynamics, decomposing abstract routing goals, and generating multi-step plans composed of multiple skills, all of which require accurate high-level reasoning during execution. 
In this paper, we propose a fully autonomous hierarchical framework for solving challenging DLO routing tasks. Given an implicit or explicit routing goal expressed in language, our framework leverages vision-language models~(VLMs) for in-context high-level reasoning to synthesize feasible plans, which are then executed by low-level skills trained via reinforcement learning. 
To improve robustness over long horizons, we further introduce a failure recovery mechanism that reorients the DLO into insertion-feasible states.
Our approach generalizes to diverse scenes involving object attributes, spatial descriptions, implicit language commands, and \myred{extended 5-clip settings}.
It achieves an overall success rate of 92\% across long-horizon routing scenarios.
Please refer to our project page: 
\url{https://icra2026-dloroute.github.io/DLORoute/}

\end{abstract}


\section{Introduction}

Deformable linear objects (DLOs) are ubiquitous in daily life and industrial applications, yet they remain challenging for robotic manipulation. Their inherent flexibility and underactuated nature, stemming from unpredictable deformation, pose significant difficulties for both action-level control and modeling of deformable dynamics. Moreover, these properties impose heightened demands on high-level task reasoning when manipulating DLOs.
For example, common routing tasks such as desk cable management require navigating cables through constrained and cluttered environments. This involves selecting appropriate entry points and accurately passing cables through holes, tubes, or clips in the correct sequence and orientation, ultimately arriving at destination sockets with suitable cable length and alignment. Similarly, industrial tasks such as automobile assembly require routing wires and signal lines along predefined paths within a vehicle frame while ensuring safety, avoiding sharp bends, and minimizing tension to prevent damage. Failures such as misalignment during insertion or undesirable DLO configurations can compromise task success, necessitating replanning and reevaluation of the entire routing process.
\begin{figure}
   
    \begin{center}
    \includegraphics[width=1.0\columnwidth]{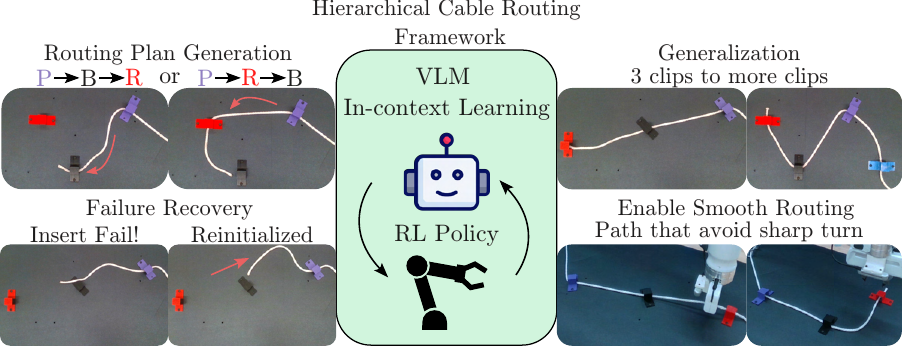} 
    \end{center}
    \vspace{-4mm}
    \caption{\textbf{Hierarchical DLO routing framework.} 
    Our framework combines high-level planning via a VLM with in-context learning and low-level control via an RL policy. The VLM generates routing plans and handles failure recovery, while the RL policy executes precise manipulation. This framework enables recovery from insertion failures through reinitialization, generalizes from three-clip to multi-clip routing, and produces smooth paths that avoid sharp turns.
    }
    \label{fig:coverpg}
    \vspace{-8mm}
\end{figure}

Previous research has explored deformable object manipulation but has struggled with limited generalization, often focusing on single-skill settings or short-horizon scenarios. Early studies addressed fundamental tasks such as insertion and pulling under frictional environments for DLOs with varying physical properties \cite{ropeinsert2024, ropepull2025}. While these works advanced low-level control, they primarily targeted short-horizon tasks involving a single goal, without addressing multi-stage, long-horizon planning. Luo et al. \cite{multistage_ucb}, for example, proposed an imitation learning approach for routing cables through clips using human demonstrations. Although effective within the training domain, their method generalizes poorly to out-of-distribution scenarios, limiting its applicability to real-world DLO routing problems.


With the advent of VLMs, robotics has begun leveraging their powerful commonsense reasoning and scene understanding capabilities. These models can provide contextual knowledge, high-level task guidance, and even recovery strategies when a robot becomes stuck or encounters failure. For instance, VLM-PC~\cite{cmnsense_leg2025} employs a VLM to guide a legged robot through unstructured environments by reasoning about progress and replanning dead-end cases, while relying on a locomotion controller for low-level actions. However, such approaches are unsuitable for DLO manipulation, where careless actions can damage the object. For example, pushing without regard to obstacles can lead to deformation or breakage. Thus, in addition to semantic reasoning for high-level planning, safe and reliable low-level control is essential for long-horizon DLO routing.


To address these challenges, we propose a hierarchical framework that integrates high-level planning via VLMs with reinforcement learning-based low-level control for DLO routing (Fig.~\ref{fig:coverpg}). The objective is to route a DLO through multiple clips in a specified or natural order based on a language prompt. \myred{Our framework is designed for longer-horizon tasks and generalizes from 3-clip to more challenging 4- and 5-clip settings.} We design three core low-level skills: \textbf{Insert} and \textbf{Pull} for clip routing, and \textbf{Flatten} for failure recovery. Pull and Flatten are inherently safe as they move the DLO away from the environment and can be initialized from predefined motion primitives. In contrast, insertion requires precise navigation near a clip to maximize insertion success while avoiding collisions, so we train this skill using reinforcement learning to balance safety and accuracy.


These low-level skills are provided to the VLM as in-context examples, each labeled as Insert, Pull, or Flatten, and each clip is annotated with spatial or attribute labels (e.g., color). We supply the VLM with task descriptions, skill definitions, and a scene image to produce a routing plan, including the clip order and insertion directions. During execution, the VLM receives both a full-scene image and a zoomed-in view of the current clip to infer task progress, select the next skill, and determine the target clip (Fig.~\ref{fig:realprocedure}). If repeated failures occur and no progress is made, the VLM recognizes the failure and triggers the Flatten skill to recover from the stuck state and resume the routing sequence.

The primary contributions of this paper are as follows:
\begin{itemize}
   \item We introduce a hierarchical framework for long-horizon DLO routing through arbitrarily arranged clips with diverse attributes, integrating low-level skills learned via reinforcement learning with high-level in-context reasoning enabled by a vision-language model.
   \item We propose a failure-aware mechanism in which the vision-language model detects execution failures, reasons about their causes, and replans accordingly, substantially improving the overall task success rate.
   \item We demonstrate that our approach achieves high performance in long-horizon DLO routing, robustly handling diverse DLOs and clip configurations with varying poses and 3-clip and multi-clip settings.
\end{itemize}

\begin{figure*}
    \begin{center}
    \includegraphics[width=0.9\textwidth]{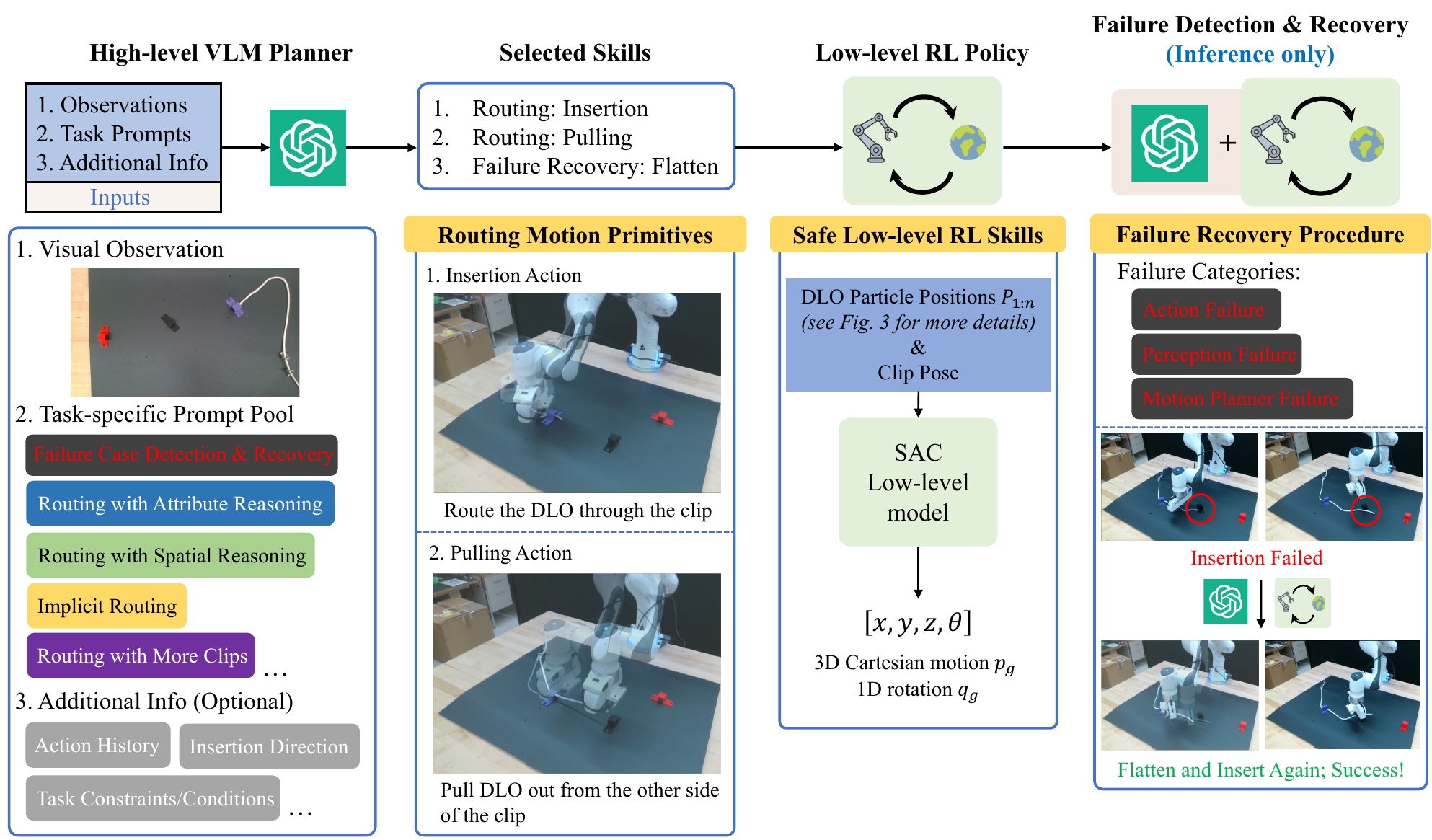} 
    \end{center}
    \vspace{-4mm}
    \caption{\textbf{Pipeline of the proposed hierarchical DLO routing framework.} The high-level VLM-based planner processes top-down scene images, task prompts, and auxiliary information to select appropriate skills, including routing (insertion and pulling) and failure recovery (flattening). Insertion is performed by a safe, low-level RL–based parameterized motion primitive for precise manipulation. A failure detection and recovery module (inference only) monitors execution to identify failures and triggers replanning, enabling robust recovery and successful long-horizon routing. 
    Note that the planner receives only top-down views.}
    \label{fig:pipeline}
    \vspace{-5mm}
\end{figure*}

\section{Related Works}

\subsection{Deformable Manipulation}

DLO manipulation has been explored in various applications, such as surgical tasks~\cite{tactileinsertion}, rope shaping~\cite{entong_diff}, and cable grasping \cite{DexDLO2024}. Recent works such as~\cite{DexDLO2024, lv2023samrl} address tasks like needle threading, pulling, and shaping. However, these approaches rely on coarse geometric models (e.g., capsules), which fail to capture DLO deformation accurately and cannot reliably predict object states.

To overcome these limitations, other research has focused on learning DLO dynamics and deformable–rigid contact interactions. This enables force-aware manipulation and improves generalization to deformable objects with varying physical properties~\cite{zhang2024adaptigraph, vandermerwe2025deformrigidcontact, ropeinsert2024}.
Simulation platforms such as SoftGym~\cite{corl2020softgym} and IsaacLab~\cite{issaclab2023, wang2025dexgarmentlab} provide more accurate deformable models and integrate them with reinforcement learning environments. Nevertheless, these works primarily focus on short-horizon tasks, leaving long-horizon DLO manipulation relatively underexplored.

One prior study proposes a hierarchical framework for cable routing by twisting a DLO through three clips~\cite{multistage_ucb}. However, their high-level controller is trained with imitation learning, which relies on limited demonstrations and struggles to generalize, leading to a substantial performance drop when extended to a four-clip setting. In contrast, our framework leverages robust low-level RL policies that accurately capture DLO dynamics with a high-level planner powered by a VLM. This combination addresses the generalization bottleneck and enables scalable, long-horizon DLO manipulation across diverse task configurations.

\subsection{Vision-language Models for Long-horizon Planning}
Vision-language models have demonstrated impressive performance in long-horizon planning, mainly by generating high-level task plans~\cite{jiang2024roboexp, goldberg2025bloxnet, llm_behavtree2025, HuangEtAl2025} or through reward generations~\cite{keyptvlm2025}. 
Our proposed approach differs from prior work in that the high-level planner is explicitly designed to interleave with low-level skill execution. 
Recent studies~\cite{hilmthome2019, li2025hamster, rarl2025, CurricuLLM2025, lee2025molmoact} have also explored combining high-level reasoning with VLMs while simultaneously learning low-level control policies. 
However, their efforts primarily target rigid-object manipulation, limiting their applicability to the more challenging domain of DLO manipulation.

\subsection{Failure Cases Detection and Recovery}
Enabling robots to autonomously detect and recover from failures has become an important challenge in the robotics community~\cite{elhafsi2023semantic, liu2023reflect, duan2024aha, agia2024unpacking, du2023vision, huang2025fail2progress}, as it's a key step toward lifelong learning during deployment in diverse real-world environments. 
However, existing work has either concentrated solely on failure detection~\cite{elhafsi2023semantic, duan2024aha, du2023vision, agia2024unpacking} or on the manipulation of rigid objects~\cite{liu2023reflect, huang2025fail2progress}. 
In contrast, our approach addresses failure reasoning and recovery within a hierarchical system, where high-level planning interleaves with low-level policies, with an emphasis on the challenging domain of DLO manipulation.

\section{Method}
\subsection{Problem Formulation}
The long-horizon DLO routing problem can be divided into two subtasks: (1) a high-level planner that reasons over scene knowledge and makes sequential decisions, and (2) a set of low-level skills that interact with the environment to perform DLO manipulation.

The planner is given an image of a scene $I_{scene}$ with three or more clips placed at different orientations and locations, along with a text description indicating the order in which it should route the DLO through. As shown in Fig. \ref{fig:pipeline}, the planner must generate a routing plan consistent with the provided description. The clip may have different attributes like color or spatial relation. The order will describe the color attributes, spatial relation, or an implicit order. We also consider \myred{extended multi-clip settings (up to 5 clips)} to evaluate long-horizon generalization.
In order to achieve the routing goal, a skill set containing possible skills is required. The planner should decide the next low-level skill to execute, the target clips to route through, and reason about task completion, given the scene image and history plan. Details about the planner are in Section \ref{hilevel_method}
Additionally, the planner must decide when to stop execution by reasoning over task progress and comparing it with the goal specified in the initial instruction.

\begin{figure}[t]
\vspace{2mm}
    \begin{center}
    \includegraphics[width=0.8\columnwidth]{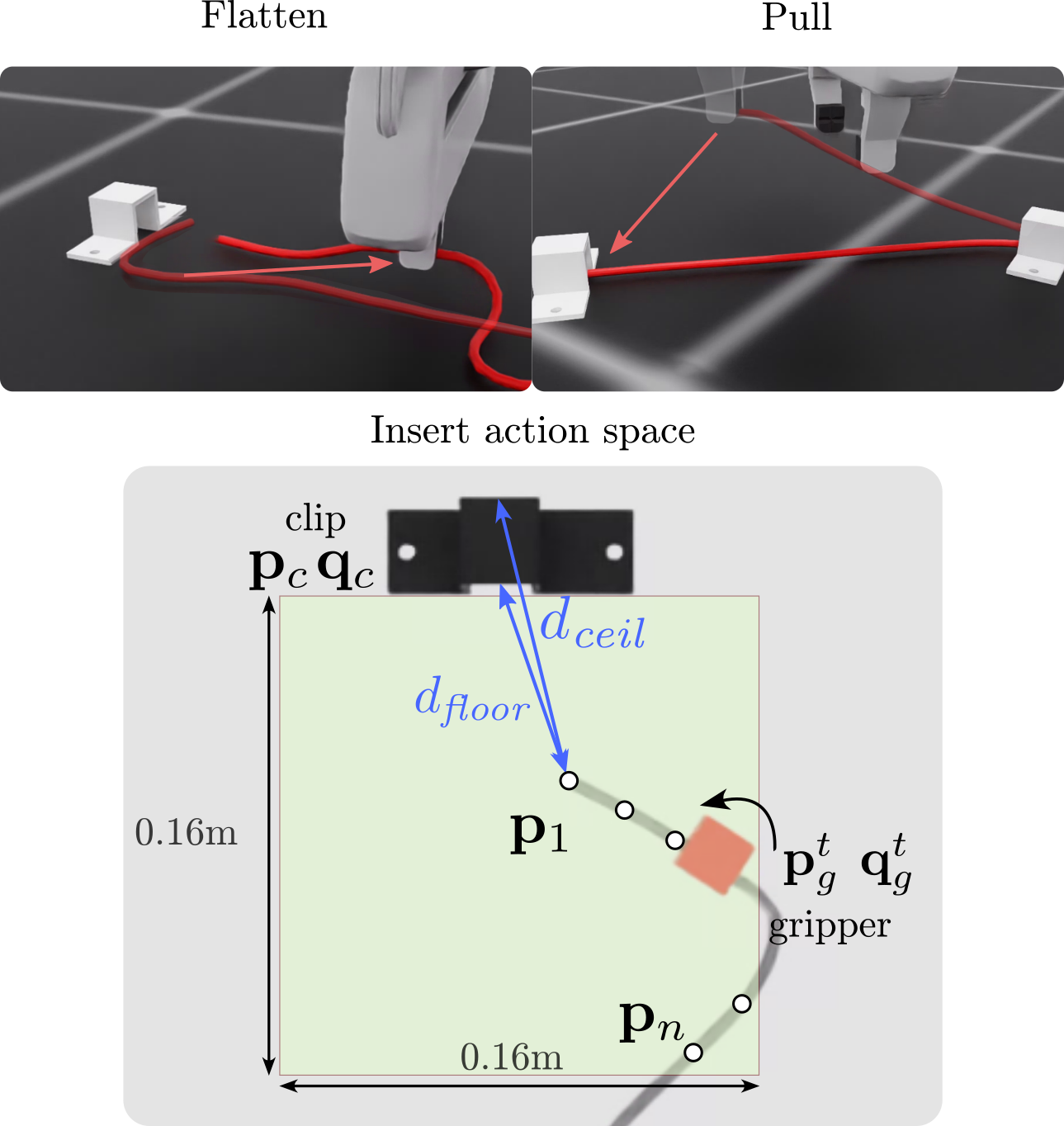} 
    \end{center}
    \vspace{-5mm}
    \caption{\textbf{Illustration of the low-level action space for DLO routing skills.} The primitive set includes Flatten (top left) and Pull actions (top right), and the Insert action (bottom). For insertion skill, the gripper (in red) operates within a $0.16m \times 0.16m$ space with orientation ($\textbf{p}_g^t$, $\textbf{q}_g^t$) conditioned on the clip (in black) state ($\textbf{p}_{c}$, $\textbf{q}_{c}$). The DLO is represented by particles $\{\textbf{p}_1, ..., \textbf{p}_n\}$.
    }
    \label{fig:act_space}
    \vspace{-4mm}
\end{figure}

Our low-level skills can be formulated as a Markov Decision Process (MDP) $(\mathbf{S}, \mathbf{A}, r, \gamma)$. The state $\mathbf{S}$ consists of the pose of the clip, and the positions $\mathbf{p}_{1:n}$ of a DLO, represented by $n$ particles as shown in Fig. \ref{fig:act_space}. The action space $\mathbf{A}$ includes 3D Cartesian motion $\mathbf{p}_g^t$, 1D rotation $\mathbf{q}_g^t$ of the robot gripper at time step $t$.
Section \ref{lowlevel_method} details the low-level skills setup and training of an RL policy for the DLO insertion skill.

\subsection{Training Low-level Skills}
\label{lowlevel_method}
Low-level skills should robustly achieve their own manipulation goal. We designed three low-level skills available to the planners: Insert, Pull, Flatten, shown in Fig. \ref{fig:act_space}. Each of them requires a specified target clip. 
Among these, insertion is the most crucial for DLO routing, as it directly accomplishes the task objective. Successful insertion must adapt to the dynamics of deformable objects and be reactive during execution in the contact-sensitive regions.
In contrast, pulling and flattening serve as auxiliary skills that adjust the DLO configuration to create more favorable conditions for insertion.
Additionally, pulling and flattening move the DLO away from the clips, whereas insertion requires precise motion close to the clip and must achieve a high success rate to minimize repeated trials.
Thus, for pulling and flattening, we employ a predefined motion primitive, while insertion is a parametrized motion primitive with its parameters optimized by a reinforcement learning agent trained in IsaacSim \cite{issaclab2023}. This allows insertion to leverage adaptive, learned behaviors for complex contact scenarios, while auxiliary skills remain simple and efficient.

We design the insertion motion primitive as shown in Fig.~\ref{fig:act_space}. First, the insertion primitive will grasp the corresponding sampled DLO segment $\mathbf{p}_i$ given a predicted grasping index $i$. After grasping, the robot follows a waypoint trajectory, with via points defined by gripper poses $(\mathbf{q}_g^t, \mathbf{p}_g^t)$ at each time step $ t$. 
\myred{The RL policy observation consists of particle-based DLO state $\mathbf{p}_{1:n}$, the clip pose and scale, and a binary indicator $rope\_in$ indicating partial DLO insertion.}
The RL policy learns to iteratively improve the insertion condition and can be executed for multiple timesteps, enabling closed-loop control and maximizing the insertion success. 

We employ Soft Actor-Critic (SAC) \cite{sac2018} for training, with multi-layer perceptron (MLP) networks for both the actor and the critic. The reward function $r$ is defined as follows:
\begin{align}
    r = &0.5 (\mathit{rope\_in}+\mathit{rope\_out}) + \beta(\mathit{collide})\\
        &+\gamma r_{hor} + r_\mathit{dist} + \eta r_{flat}
    \label{eq:reward:main}
\end{align}
where $rope\_in$ and $rope\_out$ are binary indicators denoting a halfway-through insertion or full passage through the clip, following a reward design similar to~\cite{ropeinsert2024}. The $collide$ indicates whether the robot is colliding with the clips or not. The $r_{hor}$ is the current episode length and penalizes long episodes and encourages efficient insertions with fewer time steps. The $r_\mathit{dist}$ is a stage-specific reward, while $r_{flat}$ encourages flattening the DLO to facilitate subsequent insertions. The $\beta, \gamma, \eta$ are hyperparameters to be tuned. The $r_\mathit{dist}$ is defined as follows:
\begin{equation}
    r_\mathit{dist} = 
    \begin{cases}
    10 \cdot d_\mathit{floor}\text{ }\text{ if $\mathit{rope\_in}$ } \text{ and not } \text{$\mathit{rope\_out}$}\\
    20 \cdot d_\mathit{ceil} \,\,\,\,\, \text{ if $\mathit{rope\_out}$} \\
     {1}_\mathcal{R}(\boldsymbol{p}_1)/(4+80\cdot d_{floor})~\text{otherwise}\\
    \end{cases}
    \label{eq:reward:dist}
\end{equation}
where $d_\mathit{floor}$ and $d_\mathit{ceil}$ are distances from the first DLO particle $\mathbf{p}_1$ to the clip center at the floor and ceiling (Fig.~\ref{fig:act_space}), $\mathbf{1}_{\mathcal{R}}(\mathbf{p}_1)$ refers to an indicator function whether the DLO head $\mathbf{p}_1$ lies within a predefined region ahead of the clip, discouraging DLO head $\mathbf{p}_1$ from being moved out of the insertion-possible region.
The $r_{flat}$ encourages the front segment of the DLO to stay straight in front of the clip, keeping insertion in a straightforward state as follows:
\begin{equation}
    r_{flat} = 1/(1+\frac{1000}{3}\sum_{i=0}^3 ||\mathbf{p}_{i+3}-\mathbf{p}_i||_y).
    \label{eq:reward:flat}\\
\end{equation}
To encourage a faster convergence, we apply curriculum learning, where the RL agent initially trains under simple randomization and is exposed to more complex scenes.

\subsection{High-level In-context Learning}
\label{hilevel_method}
For a high-level VLM planner, we leverage in-context learning to adapt the robot to our task domain and output reliable low-level skill instructions. 
\myred{The VLM planner operates strictly at the high-level layer. It decides the routing order and selects among a predefined set of skills and target clips, but does not generate low-level actions. To predict the routing order, the planner takes as input a top-down scene image $I_{scene}$ and a user-specified routing instruction.}
\myred{In subsequent queries, it outputs a structured decision: selected skill, target clip, and task completion flag.}
To achieve this, we instruct VLM to perform 1) Task progress reasoning, 2) low-level skills reasoning. 
Along with the user's text, we provide a scene description that includes the attributes of the DLO, clips, and environment to help improve scene understanding. 
Since the locations of the DLO front segments are crucial for the routing with DLO head manipulation, we ask the planner to track the DLO head throughout the task and to recognize occlusions caused by clips when the DLO has already passed through them. This enables more accurate reasoning about task progress, target clip selection, and completion status. 
To further enhance decision making, we employ chain-of-thought (CoT) prompting~\cite{cot24} to ask the planner to analyze the zoom-in view, thereby determining DLO–clip interaction before predicting the next skill and target clip.
\myred{Skill reasoning helps the planner distinguish between available skills and make precise selections. We provide the VLM with structured definitions for each skill (Insert, Pull, Flatten), including their functional objectives, usage conditions, and counter-examples specifying when they should not be invoked. For instance, insertion is conditioned on the rope head being close to and roughly aligned with the clip opening. We provide two in-context examples for insertion and pulling scenarios, each consisting of an image and its corresponding skill, similar to the leftmost examples in Figure \ref{fig:realprocedure}. The prompt also specifies criteria for insertion success and task success.}

\subsection{Hierarchical Planning \& Failure Recovery}

During the DLO routing rollout with only insertion and pulling skills, we observed that when pulling the DLO toward the next clip, the DLO head may twist or drift far away from the clip opening. Under such states, insertion attempts will consistently fail. To address this problem, we introduce a failure recovery mechanism. The recovery module consists of a low-level rescue skill, Flatten, a simple yet effective failure-recovery skill that reintializes the DLO to an insertion-possible state.
In addition to providing the definition and usage example of flattening skill, we set a counter of consecutive insertion attempts and feed this status to the planner. We then impose a step limit to prevent getting stuck by infinitely executing the same skill and incorporating this constraint into the CoT prompting.
Failure detection and recovery are automatically predicted by the planner from observations and history, without human intervention.
\myred{With this design, even if the DLO head aligns with the clip’s long axis, making insertion difficult, the planner can invoke recovery to reorient the DLO and resume insertion.}


\begin{figure}
    \vspace{2mm}
    \begin{center}
    \includegraphics[width=1.0\columnwidth]{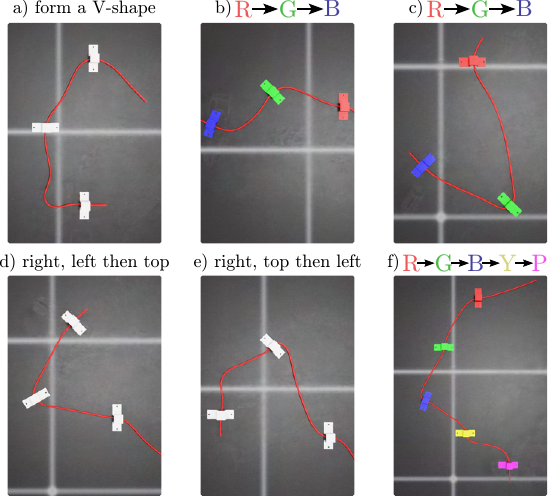} 
    \end{center}
    \vspace{-5mm}
    \caption{\textbf{Long-horizon routing examples in simulation using our VLM with failure reasoning and RL policy.} (a) Implicit order forming a V-shape. (b–c) Fixed order based on color attributes (red, green, blue). (d) Fixed order determined by spatial reasoning. (e) Fixed order based on color attributes in a 5-clip arrangement.}
    \label{fig:4scene}
    \vspace{-4mm}
\end{figure}

\section{Experiments}

We perform the experiments to answer the following research questions: 
\textbf{Q1}: Can our approach with a hierarchical framework excel at long-horizon deformable manipulation tasks? 
\textbf{Q2}: Does failure reasoning boost the performance during long-horizon reasoning? 
\textbf{Q3}: Would our approach generalize to unseen scenarios~(e.g., more clips)? 
\textbf{Q4}: Does our approach outperform baselines using a heuristic policy as the low-level policy? 
In this paper, we train our low-level RL policy in IsaacSim~\cite{issaclab2023} and GarmentLab~\cite{lu2024garmentlab} and use GPT-5 \cite{openai2025gpt5} with low reasoning effort as the VLMs planner for our experiments.

\subsection{Experiment Setup and Evaluation Metrics}
In this section, we present experiments on closed-loop RL policy training for DLO Insertion, VLM planner setup, and a comparative performance analysis against several baselines.
\subsubsection{Low-level Skills Setup}
Low-level skills consist of the Insert, Pull, and Flatten, with the insertion skill designed as a parameterized motion primitive and trained using reinforcement learning. The Pull and Flatten primitives have predefined motion if the target clip is determined as shown in Fig.~\ref{fig:act_space}.

\begin{table}[tbh]
\begin{center}
\scalebox{1.0}{
\begin{tabular}{ccc} 
\toprule
Metrics &  Heuristic Policy& RL Policy (Ours)\\ 
\midrule
Success Rate (\%) $\uparrow$ & 45 &\textbf{87}\\
Avg. dis. (cm) $\uparrow$  & 1.24& \textbf{2.59}\\
Episode Length & 2& 3.86 \\
\bottomrule
\end{tabular}
}
\end{center}
\vspace{-4mm}
\caption{\textbf{Simulation result for rope insertion primitive.}}
\label{tab:sim_rl}
\vspace{-3mm}
\end{table}

\begin{table*}
\vspace{2mm}
\begin{center}
\centering
\resizebox{\textwidth}{!}{%
\begin{tabular}{c|c|ccc|ccc|ccc|ccc}
\hline
\multirow{2}{*}{High Level} & \multirow{2}{*}{Low Level} 
& \multicolumn{3}{c|}{Implicit} 
& \multicolumn{3}{c|}{Fixed (Spatial)} 
& \multicolumn{3}{c|}{Fixed (Attr.)} 
& \multicolumn{3}{c}{4/5-Clip} \\
\cline{3-14}
& 
& SR $\uparrow$ & Clips $\uparrow$ & Epi 
& SR $\uparrow$ & Clips $\uparrow$ & Epi
& SR $\uparrow$ & Clips $\uparrow$ & Epi
& SR $\uparrow$ & Clips $\uparrow$ & Epi \\
\hline

VLM w/ Failure Reasoning & Heuristic Policy 
& \myred{13} & 1.3 & 8.0
& \myred{7} & 0.7 & 6.9
& \myred{0} & 0.9 & 7.3
& \myred{7} & 1.3 & 9.1 \\
\hline

\myred{VLM w/o Failure Reasoning} & RL Policy
& \myred{47} & 2.0 & 16.2
& \myred{7} & 1.3 & 22.7
& \myred{7} & 1.2 & 23.3
& \myred{47} & 3.1 & 28.4 \\

\hline

Fixed Order & RL Policy
& \myred{53} & 2.1 & 9.7
& \myred{13} & 1.3 & 8.7
& \myred{13} & 1.3 & 8.9
& \myred{60} & 3.4 & 14.6 \\
\hline

Symbolic Planner & RL Policy
& \myred{68} & 2.3 & 10.0
& \myred{53} & 2.1 & 11.5
& \myred{53} & 2.1 & 12.0
& \myred{\textbf{100}} & \textbf{4.3} & 21.5 \\

\hline

VLM w/ Failure Reasoning & RL Policy
& \myred{\textbf{80}} & \textbf{2.7} & 13.9
& \myred{\textbf{93}} & \textbf{2.9} & 18.3
& \myred{\textbf{93}} & \textbf{2.9} & 17.7
& \myred{\textbf{100}} & \textbf{4.3} & 20.7 \\

\hline
\end{tabular}
}
\end{center}

\vspace{-4mm}
\caption{\textbf{Simulation results for long-horizon DLO routing.} 
SR = success rate (\%), Clips = average number of clips inserted, Epi = average episode length. Higher SR and Clips indicate better performance.}
\label{tab:simeval}
\vspace{-5mm}
\end{table*}

In both simulation and real-world experiments, we constrain the insertion manipulation trajectories to a 2D plane and 1D rotation with respect to the z-axis, resulting in a 3-dimensional waypoint for a motion primitive. One motion primitive consists of two via points ($\mathbf{p}_{g}^{T} \in \mathbb{R}^2$, $\alpha_{g}^T \in \mathbb{R}$, $T \in {1,2}$) and one indexed grasping location that grasp at DLO particle $\mathbf{p}_i$, totaling seven parameters for the insertion motion primitive. For the reward function, we use $\beta=-2$, $\gamma=-0.001$, and $\eta=0.5$. 

In the simulation, we randomized the DLO position and angle within a 10cm$\times$5cm rectangle and between -10 $\deg$ to 10 $\deg$, \myred{with a friction coefficient of 0.5.} For curriculum learning, we randomize the clip size to control the task difficulty. The original clip has a 2.2cm opening. For the first 1600 training steps, the clip scale is randomized from 1.0 to 2.2, providing a maximum clip opening of 4.84cm wide. After the policy is well initialized, we then focus the training on a domain-specific clip scale from 0.9 to 1.5. The RL policy is trained for 6.2k steps, and the evaluation result is shown in Tab. \ref{tab:sim_rl}.

During long-horizon DLO routing tasks, the agent will encounter clips with vastly different orientations and poses. To alleviate this confusion for low-level policy, we always transform the DLO state observation back into the clip's local frame and execute the action in the local frame.

\subsubsection{Evaluation Details}
As shown in Tab. \ref{tab:sim_rl}, we evaluate the low-level policies on 100 different scenes with randomized DLO poses and clips at the original scale, \myred{matching the real-world clips used in our experiments.}
Average distance (avg. dis.) is the average signed endpoint distance $\pm d_\mathit{ceil}$. It is positive when $\mathit{rope\_out}$ and negative otherwise. The success is recorded when the signed endpoint distance $\pm d_\mathit{ceil}$ is larger than +2cm after the motion primitive finishes.

As for long-horizon DLO routing, we evaluate our baselines with four different DLO routing strategies: implicit orders, fixed orders with spatial descriptions or color attributes, and fixed orders with color attributes under \myred{4/5-clip settings, as shown in Fig. \ref{fig:4scene}. 
Each strategy is evaluated over 15 trials with different clip layouts.
} The DLO must fully pass through each clip, and the DLO head must emerge from the other side. A success case is recorded when the DLO is routed through all clips in the scene with the correct order. Average episode length means the number of pick-and-place steps executed. Average number of clips inserted shows the average number of clips inserted when the episode terminates. An episode may be terminated and be considered a failure case when 1) a collision is detected, 2) the maximum timeout has been reached, or 3) early termination with some clips remaining not fully inserted.

\subsection{Baselines}
We conduct various open-loop and closed-loop low-level insertion experiments in our simulation. For our RL policy, the horizon of manipulation could take up to 7 steps to allow the agent to fully attempt the scene. The following baselines are open-loop, where the agent executes an entire trajectory and terminates the episode:
\begin{itemize}
    \item \myred{\textbf{Heuristic Policy}: An insertion method that imitates a human expert’s strategy. The robot grasps the DLO head and places it in front of the clip, then regrasps further along the DLO and inserts it along the clip centerline.}
\end{itemize}
For long-horizon DLO routing evaluation, we employ several heuristic methods and conduct ablation studies. 
\begin{itemize}
    \item \textbf{VLM (w/o failure reasoning)}: Remove the failure reasoning and recovery description from the prompt text and remove the Flatten primitive. The VLM planner can only command insert and pull to complete the task.
    \item \textbf{Fixed order}: Given the ground truth insertion order, we adopt a heuristic method that repeatedly executes Pull and Insert actions until the task either succeeds or fails.
    \item \myred{\textbf{Heuristic policy as low-level policy}}: substitutes the low-level RL policy with the heuristic strategy which is less considerate of the environment, allowing us to evaluate performance under a simpler motion primitive.
    \item \myred{\textbf{Symbolic Planner}: The planner uses ground-truth clip poses and the rope head position to decide whether to execute Insert or Pull. The routing order is determined using a greedy strategy that selects the next closest clip. Insertion is considered successful when the rope head passes through the clip opening; otherwise, it retries insertion and triggers Flatten after three failures.} 
\end{itemize}

\subsection{Simulation Experiments}


In this section, we will address the aforementioned questions using our simulation results summarized in Table~\ref{tab:simeval}.

\begin{figure*}
    \begin{center}
    \includegraphics[width=1.0\textwidth]{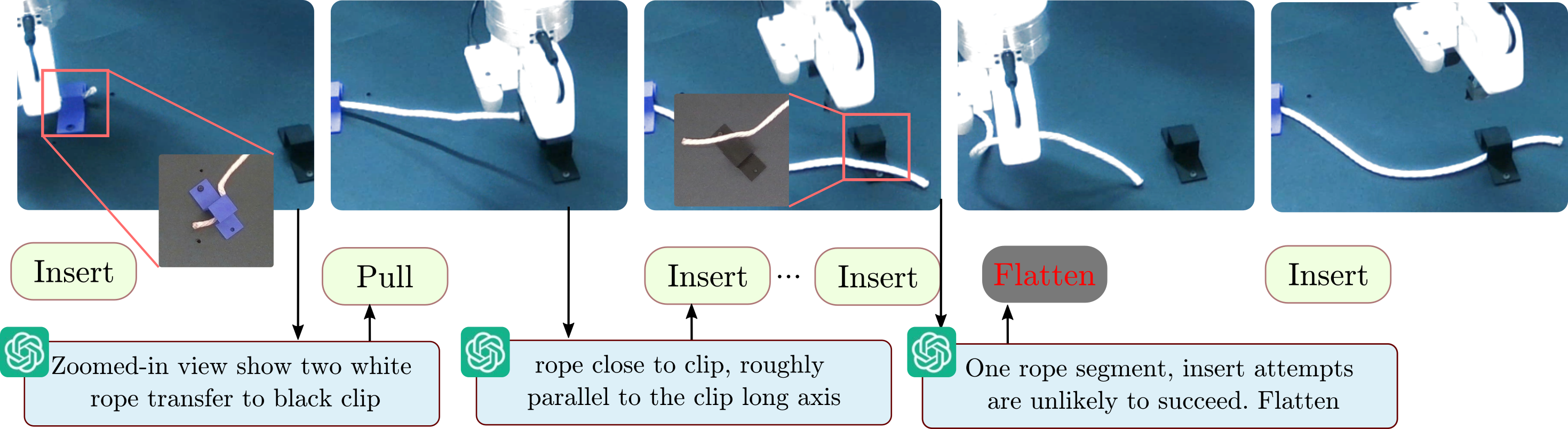} 
    \end{center}
    \vspace{-4mm}
    \caption{\textbf{Real-robot execution of the proposed hierarchical DLO routing framework.} The robot receives both a whole-scene view and a zoomed-in view centered on the current clip. During normal execution, it alternates between insertion and pulling actions. When insertion becomes unlikely due to unfavorable DLO configurations (e.g., alignment along the clip’s long axis or sliding past the clip without entering), the system detects the failure, executes a flatten action to reinitialize the DLO, and then resumes insertion successfully.}
    \label{fig:realprocedure}
    \vspace{-3mm}
\end{figure*}

\begin{figure}[t]
    \begin{center}
    \includegraphics[width=0.8\columnwidth]{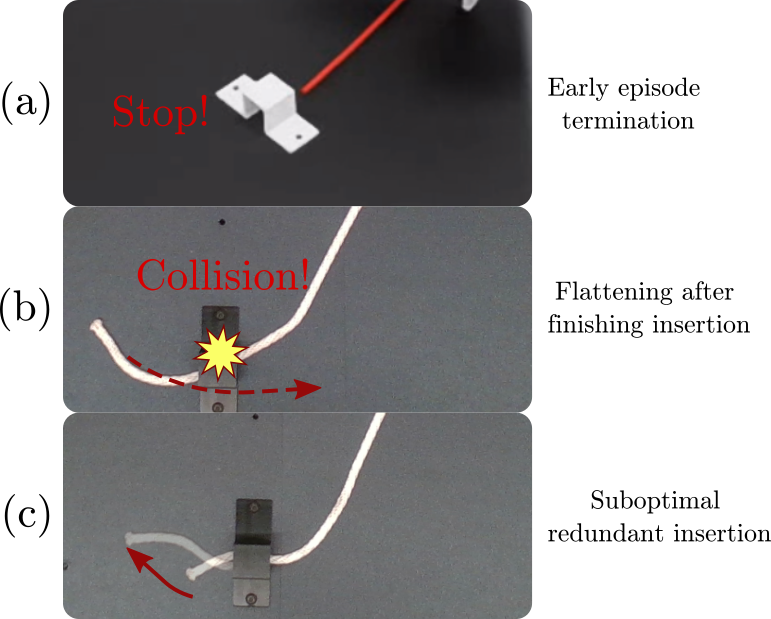} 
    \end{center}
    \vspace{-5mm}
    \caption{\textbf{Representative failure cases observed in simulation and real-world experiments}: (a) early episode termination before completing last-clip insertion, (b) unintended collision when falsely flattening after an insertion, and (c) suboptimal redundant insertion where the VLM should choose pull for a more effective plan.
    }
    \label{fig:failure}
    \vspace{-6mm}
\end{figure}

\textbf{Q1}: 
The hierarchical planning has greatly improved the performance compared with the fixed order and heuristic policy baselines. As shown in Table~\ref{tab:simeval}, our method excels across all tasks, achieving a higher success rate and a greater number of clips inserted across all evaluation settings, even under challenging conditions. While our method requires longer episode lengths, the improved robustness and task completion demonstrate that the hierarchical framework excels at long-horizon deformable manipulation tasks.
\myred{The symbolic planner baseline highlights the limitations of purely heuristic planning. Its greedy routing strategy cannot process language instructions and may select the closest clip instead of the correct target, particularly under challenging conditions such as Implicit, Fixed Spatial, Fixed Attribute. When the greedy order happens to match the correct sequence (e.g., in the 4/5-clip setting), it can achieve high success rates but still results in longer episodes because it repeatedly attempts insertion before recovery. In contrast, our method detects failures earlier and invokes the Flatten action more efficiently.} Certain failure cases for ours are shown in Fig.~\ref{fig:failure}, the high-level planner issues an early termination before the DLO has been fully inserted into the last clip.

\textbf{Q2}: Incorporating failure reasoning boosts our performance over the plain VLM policy. Across all evaluation settings, our method achieves higher success rates and more clips inserted, with the most substantial gains in all challenging long-horizon scenarios. Moreover, episodes are shorter in these cases, indicating that the agent recovers more efficiently from errors. In contrast, an agent without failure reasoning can only try several times under insertion-challenging scenarios and keep getting failure results. 
\myred{As shown in Table II, baselines without failure reasoning (e.g., the plain VLM policy and Fixed Order) perform reasonably well in the implicit setting. In this case, the clip opening angles are similar, producing mild turns between clips and forming a natural routing order. Such configurations reduce the need for failure recovery, allowing these baselines to succeed.}
However, scenarios such as Fixed Spatial and Fixed Attribute are relieved from the mild turn preset condition and contain over 90-degree turns between clips. This makes insertion much harder without recovering from failure.
These results demonstrate that failure reasoning is a key factor in boosting long-horizon reasoning performance.

\textbf{Q3}: Our approach can extend beyond the current clip setting to a more challenging \myred{4/5-clip settings}. Although this setting is extended from relatively simple routing scenarios, the added clip usually forms a large angle compared with the previous clip, making the setting much harder for simple insertion strategies. Despite this difficulty, our hierarchical framework with failure reasoning achieves a perfect success rate and consistently inserts all clips, demonstrating strong generalization to longer-horizon tasks. By contrast, both the VLM-only policy and the Fixed Order baseline suffer notable drops in success rate and efficiency. 
\myred{These results indicate that our approach scales well with the number of clips, as the high-level planner reliably predicts next skills while the low-level policy performs consistently across different clips.}

\textbf{Q4}: Comparing against the heuristic policy shows that the low-level policy plays a critical role in long-horizon deformable manipulation. While the heuristic insertion baseline quickly fails due to its lack of environmental awareness, the RL low-level policy adapts actions to task progress and scene dynamics, enabling precise and robust insertions. This distinction becomes most evident in the 4/5-Clip setting, where the heuristic policy achieves only 7\% success while our approach reaches 100\%, demonstrating that task-aware low-level skill is essential for scaling to complex scenarios.
\subsection{Real-world Robot Experiments}
We conducted real robot long-horizon routing tasks using a Franka Emika Panda robot. A wrist-mounted, calibrated Intel RealSense D415 camera captures top-down images of the tabletop scene at a resolution of $1280 \times 720$. The MoveIt planner \cite{chitta2012moveit} was used to generate continuous trajectories based on the motion primitives predicted by our model or predefined motion skills. The overall system was built using the Robot Operating System (ROS) framework \cite{quigley2009ros}, and clips and DLO are segmented using SAM2 \cite{ravi2024sam2} with an NVIDIA 5090 GPU. The reference points for DLO segmentation are initially set around the fixture and iteratively added during segmentation. 
\myred{Our RL policy uses state-based observations extracted from segmentation. During simulation training, we randomize rope configurations to capture a wide distribution of states, enabling the policy to generalize to real-world rope configurations.}
Consequently, our policy did not require additional data for fine-tuning or domain adaptation during real robot testing.
We evaluate our method on 8 clip configurations and compare our approach to the Fixed Order baseline across the 4 tasks mentioned in Tab.~\ref{tab:simeval}. We report the success rate and the average number of clips inserted, as shown in Tab.~\ref{tab:realexp}. The routing process is shown in Fig. \ref{fig:realprocedure}.
Although the success rate in real experiments is lower than in simulation, our method still achieves strong performance, reaching 62.5\% success despite challenges from calibration errors, perception noise, and sim-to-real transfer. In contrast, the Fixed Order baseline performs significantly worse, highlighting that real-world routing requires explicit failure reasoning and closed-loop decision making to handle unexpected variations and disturbances.
The common failure cases in simulation and in the real world are summarized in Fig.~\ref{fig:failure}. Faulty planning has been observed in a real experiment where the robot executes the 'flatten' command even though the DLO is already inserted into the clip, leading to a grasping failure and a collision with the clip. Another suboptimal plan observed in real and simulation is that the planner ignores the emerging DLO segment in the zoomed-in image, incorrectly concludes that insertion is incomplete, and issues another insertion action. While this behavior does not necessarily result in a collision since our RL policy can select an appropriate grasping location, it still unnecessarily prolongs the routing process and leads to a suboptimal plan.

\begin{table}[th]
\begin{center}
\scalebox{1.0}{
\begin{tabular}{ccc} 
\toprule
Metrics & Fixed Order + RL & Ours \\ 
\midrule
SR $\uparrow$ &  37.5  & \textbf{62.5}\\
Clips $\uparrow$ & 1.75  & \textbf{2.625}\\
\bottomrule
\end{tabular}
}
\end{center}
\vspace{-4mm}

\caption{\textbf{Real-world evaluation for long-horizon DLO routing}}
\label{tab:realexp}
\vspace{-5mm}
\end{table}

\section{Limitations \& Future Work}
While our long-horizon DLO routing framework provides a promising solution for DLO routing in diverse scenes, several failure cases remain unsolved. Addressing faulty or suboptimal planning is an important direction for future work. One potential approach is to fine-tune the VLM planner to improve planning stability. Another direction is to integrate new skills into the VLM planner automatically. This process could be automated by providing the planner with sequences of action images and letting it reason through the usage and definition of skills. 
Such an approach would enable large-scale skill integration and improve generalization to increasingly complex deformable manipulation tasks.

\section{Conclusion}
In this work, we presented a hierarchical framework for long-horizon DLO routing that integrates a high-level VLM planner with a key low-level RL skill. The VLM leverages in-context reasoning and failure recovery to generate reliable task plans and skill selections, while the RL insertion policy robustly handles complex DLO dynamics. Through extensive simulation and real-world experiments, we demonstrated that our approach outperforms other baselines, achieving strong generalization to \myred{4/5-clip settings}. This highlights the importance of combining high-level reasoning with task-aware low-level control for scalable DLO manipulation.

{
\bibliography{icra26ref,IEEEabrv}
\bibliographystyle{IEEEtran}
}

\end{document}